\documentclass[10pt,twocolumn,letterpaper]{article}
\usepackage{cvpr}  
\usepackage{graphicx}
\usepackage{amsmath}
\usepackage{amssymb}
\usepackage{booktabs}
\usepackage{multirow}

\usepackage{xcolor}
\usepackage{pifont}

\makeatletter
\DeclareRobustCommand\onedot{\futurelet\@let@token\@onedot}
\def\@onedot{\ifx\@let@token.\else.\null\fi\xspace}
\def\eg{\emph{e.g}\onedot} 
\def\ie{\emph{i.e}\onedot}

\makeatother

\definecolor{lightgray}{rgb}{0.9, 0.9, 0.9}
\definecolor{lgray}{rgb}{0.66, 0.66, 0.66}
\newcommand{\cmark}{\ding{51}\xspace}%
\newcommand{\xmarkg}{\textcolor{lgray}{\ding{55}}\xspace}%

\usepackage{array}
\newcommand{\PreserveBackslash}[1]{\let\temp=\\#1\let\\=\temp}
\newcolumntype{C}[1]{>{\PreserveBackslash\centering}p{#1pt}}
\newcolumntype{R}[1]{>{\PreserveBackslash\raggedleft}p{#1pt}}
\newcolumntype{L}[1]{>{\PreserveBackslash\raggedright}p{#1pt}}

\usepackage[pagebackref,breaklinks,colorlinks]{hyperref}

\usepackage[capitalize]{cleveref}
\crefname{section}{Sec.}{Secs.}
\Crefname{section}{Section}{Sections}
\Crefname{table}{Table}{Tables}
\crefname{table}{Tab.}{Tabs.}

\begin{document}

\title{Region-Aware Face Swapping}
\author{Chao Xu$^1$\thanks{Work Down during an intership at Bytedance.}~\thanks{Equal contribution.}
~ ~ Jiangning Zhang$^1$\footnotemark[2]
~ ~ Miao Hua$^2$
~ ~ Qian He$^2$
~ ~ Zili Yi$^2$
~ ~ Yong Liu$^1$\thanks{Corresponding author.} \\
\normalsize $^1$ APRIL Lab, Zhejiang University ~ ~ $^2$Bytedance \\
{\tt\small \{21832066, 186368\}@zju.edu.cn, yongliu@iipc.zju.edu.cn} \\
{\tt\small huamiao@bytedance.com, 1988heqian@163.com, 527994111@qq.com}
}

\twocolumn[{%
\renewcommand\twocolumn[1][]{#1}
\vspace{-2.5em}
\maketitle
\vspace{-2.5em}
\begin{center}
\centering
\captionsetup{type=figure}
      \includegraphics[width=1\textwidth]{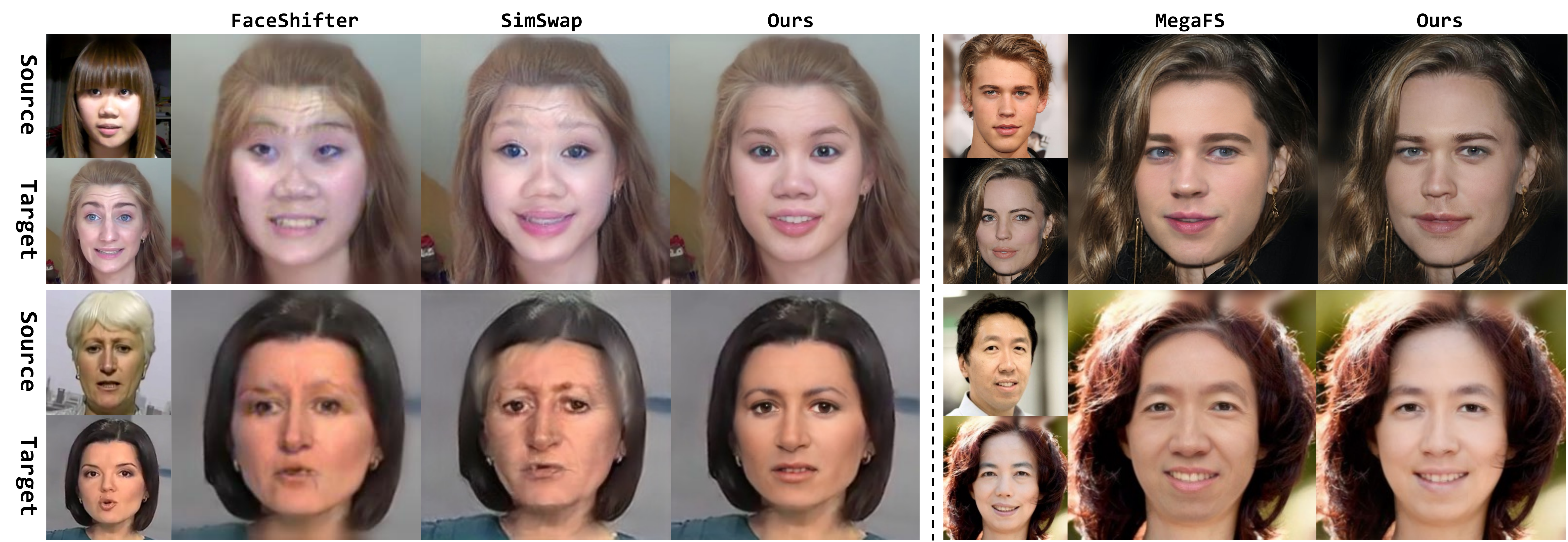}
        \caption[figure]{\textbf{Comparison with SOTA methods on challenging situations.} \textbf{Left part} shows the results of attribute-distinct cases, \eg, bangs and the white hair from the source identity, where our method is significantly better than FaceShifter~\cite{li2019faceshifter} and SimSwap~\cite{chen2020simswap} with higher quality, better identity-consistency, and fewer artifacts. \textbf{Right part} shows high-resolution results of light-changing and in-the-wild situations with SOTA MegaFS~\cite{zhu2021one}, which exists artifacts around face contour and light while our method could better preserve attributes of the target face. Images are from official attached results or released codes for fair comparisons. Please zoom in for more details.
        \label{fig:intro}}
\end{center}
}]

\maketitle
\begin{abstract}
    \vspace{-1.1em}
   This paper presents a novel Region-Aware Face Swapping (RAFSwap) network to achieve identity-consistent harmonious high-resolution face generation in a local-global manner: \textbf{1)} Local Facial Region-Aware (FRA) branch augments local identity-relevant features by introducing the Transformer to effectively model misaligned cross-scale semantic interaction. \textbf{2)} Global Source Feature-Adaptive (SFA) branch further complements global identity-relevant cues for generating identity-consistent swapped faces. 
   Besides, we propose a \textit{Face Mask Predictor} (FMP) module incorporated with StyleGAN2 to predict identity-relevant soft facial masks in an unsupervised manner that is more practical for generating harmonious high-resolution faces. 
   Abundant experiments qualitatively and quantitatively demonstrate the superiority of our method for generating more identity-consistent high-resolution swapped faces over SOTA methods, \eg, obtaining 96.70 ID retrieval that outperforms SOTA MegaFS by 5.87$\uparrow$.
   {\let\thefootnote\relax\footnotetext{$^{*}$ Work done during an internship at Bytedance.}}
   {\let\thefootnote\relax\footnotetext{$^{\dag}$ Equal contribution.}}
   {\let\thefootnote\relax\footnotetext{$^{\ddagger}$ Corresponding Author.}}
   
\end{abstract}
\section{Introduction}
\label{sec:intro}
\vspace{-0.5em}
Face swapping aims at transferring the identity of the source identity to the target identity while keeping the identity-irrelevant attributes of the target face unchanged, which has attracted widespread attention in the film industry and computer games. Recently, many researchers have achieved significant progress in face swapping, especially designing inversion-based methods to generate high-resolution face images. However, \textbf{there are two continuously critical issues}: \emph{1) How to maintain identity consistency with the source identity, including local and global facial details.} Almost all current methods~\cite{li2019faceshifter, chen2020simswap} perform feature interaction only on global feature representation without modeling identity-relevant local regions, \eg, lips, nose, brows, and eyes, which will limit the model's ability to express identity consistency. \emph{2) How to generate high-resolution swapped faces while keeping the identity-irrelevant details consistent with the target face under the GAN inversion framework, \eg, background and occlusions.} Recent works~\cite{zhu2021one, yang2021shapeediter} exploit the StyleGAN2~\cite{karras2020analyzing} as the powerful decoder but fail to maintain the consistency of the identity-irrelevant attributes of the target face. In this paper, we are dedicated to solving both the above problems.

Recent works~\cite{li2019faceshifter, li2021faceinpainter, xu2021facecontroller, chen2020simswap, wang2021hififace} regard face swapping as a style transfer task that employs global AdaIN~\cite{huang2017arbitrary} to transfer the identity information of the source face into the target face. However, the identity vector produced by the face recognition network is naturally not well-disentangled, which inevitably includes some identity-irrelevant information of the source face, \eg, background, light distribution, and hairstyle. This wrong information will be further injected into the target feature in a global manner via AdaIN, resulting in low-quality generation results. As shown in the left part of Fig.~\ref{fig:intro}, recent AdaIN-based methods cannot preserve the source identity well, in which generated faces contain excessive information of the source face in some challenging situations, \eg, bangs and white hair. 
To better preserve the identity consistency of the generated face, we explicitly model the local facial features besides global representation to perform feature interaction more finely, which also excludes the influence of the identity-irrelevant area of the source face at the same time. In this way, our method is well competent for the above challenges, as depicted in the fourth column of Fig.~\ref{fig:intro}. Specifically, we design two parallel branches to process different fine-grained information: \textbf{1)} local \textit{Facial Region-Aware} (FRA) branch to model identity-relevant feature interaction between source and target faces, which employs a \textit{Region-Aware Identity Tokenizer} (RAT), transformer layers~\cite{vaswani2017attention}, and a \textit{Region-Aware Identity Projector} (RAP) to realize misaligned cross-scale semantic interaction, \ie, lips, nose,
brows, and eyes. \textbf{2)} global \textit{Source Feature-Adaptive} (SFA) branch to complement global identity-relevant cues, \eg, skin wrinkle, for more identity-consistent results. Details can be found in following Sec.~\ref{sec:FRA} and \ref{sec:SFA}.

To achieve high-resolution face generation for more practical application, we adopt GAN inversion framework~\cite{richardson2021encoding, tov2021designing} similar to recent face swapping works~\cite{zhu2021one, yang2021shapeediter}. But these methods introduce a fatal problem of failing to preserve the background and occlusions, because vector-conditioned progressive generation will inevitably change identity-irrelevant regions. Recent MegaFS~\cite{zhu2021one} blends the high-resolution result to the target face by the pre-existing face mask in a post-processing way, while HifiFace~\cite{wang2021hififace} learns to predict face masks in a supervised manner that restricts applications. These methods must rely on ground truth face masks and usually produce artifacts around facial contours, as shown in the right part of Fig~\ref{fig:intro}. Differently, considering that pre-trained StyleGAN2~\cite{karras2020analyzing} encapsulate rich facial semantic prior, we design a \textit{Face Mask Predictor} (FMP) to predict identity-relevant soft facial mask in an unsupervised manner, \ie, without using specific mask supervision. In this way, our model achieves harmonious high-resolution face generation that keeps identity-irrelevant attributes consistent with the target face. 
In summary, we make the following three contributions:
\begin{itemize}
\item We propose a novel Region-Aware Face Swapping (RAFSwap) network, which consists of a novel \textit{FRA} branch to augment local identity-relevant features by introducing the Transformer to effectively model misaligned cross-scale semantic interaction, and a novel \textit{SFA} branch to further complement global identity-relevant cues for generating identity-consistent swapped faces.
\item We propose a \textit{FMP} module incorporated with StyleGAN2 to predict identity-relevant soft facial masks in an unsupervised manner that is more practical.
\item Abundant experiments qualitatively and quantitatively demonstrate the superiority of our method for generating more identity-consistent high-resolution swapped faces over SOTA methods.
\end{itemize}

\begin{figure*}[t!]
	\centering
	\includegraphics[width=0.98\textwidth]{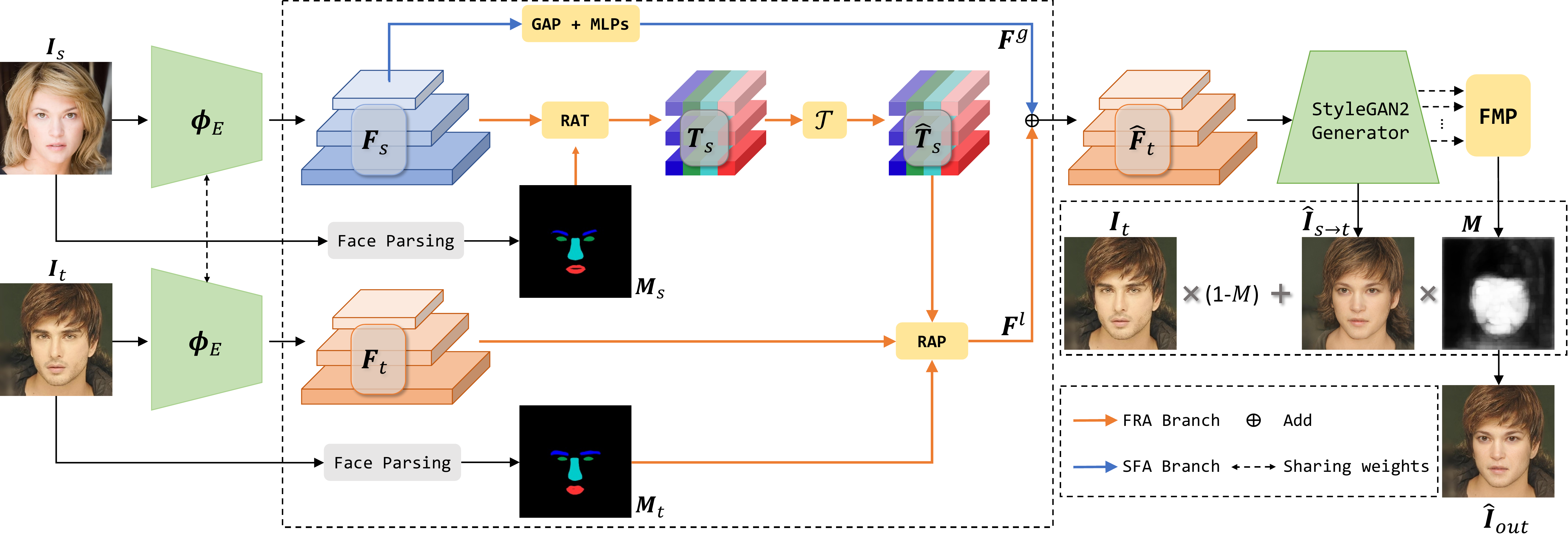}
	\caption{\textbf{Overview of the proposed RAFSwap.} 
	The source face $\boldsymbol{I}_{s}$ and target face $\boldsymbol{I}_{t}$ firstly go through a weight-sharing hierarchical face encoder $\boldsymbol{\phi}_{E}$ and a pre-trained face parsing model to obtain hierarchical features, \ie, $\boldsymbol{F}_{s}$ and $\boldsymbol{F}_{t}$, and corresponding semantic labels, \ie, $\boldsymbol{M}_{s}$ and $\boldsymbol{M}_{t}$, respectively. Then local \textit{Facial Region-Aware} branch in orange and global \textit{Source Feature-Adaptive} branch in blue are employed to integrate identity information of $\boldsymbol{I}_{s}$ with target attributes of $\boldsymbol{I}_{t}$ in a local-global manner, obtaining $\boldsymbol{F}^{l}$ and $\boldsymbol{F}^{g}$. The fused hierarchical feature $\hat{\boldsymbol{F}}_t$ is mapped into different fine-grained vectors to control the target face generation process by a StyleGAN2 generator. The \textit{Face Mask Predictor} utilize StyleGAN2 feature maps to produce the soft face mask $\boldsymbol{M}$ simultaneously. $\hat{\boldsymbol{I}}_{s \rightarrow t}$ is blended to the target face $\boldsymbol{I}_{t}$ by $\boldsymbol{M}$ to obtain final swapped face $\hat{\boldsymbol{I}}_{out}$. 
    $\hat{}$ denotes the generated face instead of the real face.
	} 
	\label{fig:pipeline}
\end{figure*}

\section{Related Work}
\subsection{GAN Inversion} GAN inversion is a task that the latent code from which the well-trained GAN could most accurately reconstruct the original input images. 
Generally, works~\cite{abdal2019image2stylegan, abdal2020image2stylegan++} directly optimize the latent vector to minimize the errors for the given images. These methods could achieve high reconstruction quality, but they are time-consuming. Subsequently, recent methods employ an encoder to map the given image to the latent space end-to-end. Specifically, pSp and GHF~\cite{richardson2021encoding, xu2021generative} embed real images into a series of style vectors that are fed into a pre-trained StyleGAN2 generator. e4e~\cite{tov2021designing} designs the encoder that generates a single base style code and a series of offset vectors to yield the final style codes. 

\subsection{Face Swapping} Face swapping aims to change the facial identity but to keep other facial attributes constant. Early efforts~\cite{blanz2004exchanging, cheng20093d, bitouk2008face} focus on 3D-based methods, but it requires manual interaction and cannot preserve the target expression. To address this limitation, Face2Face~\cite{thies2016face2face} fits a 3D morphable model (3DMM) to both the source and target faces. Nirkin \emph{et al.}~\cite{nirkin2018face} combine 3DMM and face segmentation model to achieve robust face swapping under unprecedented conditions. Besides, with the popularity of GANs~\cite{goodfellow2014generative}, learning-based methods have enabled significant progress in face swapping. DeepFakes~\cite{perov2020deepfacelab} trains an Encoder-Decoder architecture for two specific identities but lacks generalization ability.
Some works follow the disentanglement paradigm. IPGAN~\cite{bao2018towards} disentangles the identity from the source face and attributes from the target face separately and recombine them for identity preserving face synthesis. FaceShifter~\cite{li2019faceshifter} adaptively integrates identity and attribute embeddings with the attentional way. FaceInpainter~\cite{li2021faceinpainter} adopts 3D priors, texture code, and identity code for explicitly disentanglement. Recently, MegaFS~\cite{zhu2021one} first exploits StyleGAN2 as the decoder for high-resolution face swapping. However, the above referential methods struggle to generate highly identity-consistent faces due to the global feature fusion.

\subsection{Feature Fusion} Feature fusion is an important process in face swapping. Most previous works~\cite{wang2021hififace, xu2021facecontroller, chen2020simswap, li2021faceinpainter, li2019faceshifter, zeng2020realistic} are inspired by style transfer methods. They employ AdaIN~\cite{huang2017arbitrary} to inject the identity vector into the target face to generate swapped face. Besides, MegaFS~\cite{zhu2021one} proposes FTM to control multiple attributes of identity information, while other methods~\cite{yang2021shapeediter, nitzan2020face, ngo2020unified, zhang2020freenet} barely concatenate identity and attribute vectors. However, such global operations do not model the crucial local features interaction. More recently, the attention structure is playing a pivotal role in feature enhancement and interaction in NLP~\cite{vaswani2017attention, devlin2018bert, zhang2021analogous} and CV~\cite{liu2021swin, wu2020visual}. Based on the attention mechanism, we design our RAT and RAP for feature fusion, which fully fuses the local and global identity-relevant features of the source face while preserving the attributes of the target face.

\section{Method}
In this paper, a novel RAFSwap is proposed to generate high-resolution and identity-consistent swapped face images. Our method is built on a GAN inversion framework, pSp~\cite{richardson2021encoding}. As depicted in Fig.~\ref{fig:pipeline}, we first send source face $\boldsymbol{I}_{s}$ and target face $\boldsymbol{I}_{t}$ to a Hierarchical Face Encoder $\boldsymbol{\phi}_{E}$ to extract hierarchical features $\boldsymbol{F}_{s}=\left\{\boldsymbol{F}_s^{0}, \boldsymbol{F}_s^{1}, \boldsymbol{F}_s^{2}\right\}$ and $\boldsymbol{F}_{t}=\left\{\boldsymbol{F}_t^{0}, \boldsymbol{F}_t^{1}, \boldsymbol{F}_t^{2}\right\}$. Superscripts $0, 1, 2$ represent small, medium, and large scale, respectively. All feature maps are mapped to 512 channels. In the meantime, the semantic labels $\boldsymbol{M}_{s}$ and $\boldsymbol{M}_{t}$ of the source and target face on lips, nose, brows, and eyes areas are extracted by BiseNet~\cite{yu2018bisenet}. 
Second, FRA and SFA are employed to extract local and global discriminative identity features of the source face $\boldsymbol{I}_s$, obtaining $\boldsymbol{F}^{l}$ and $\boldsymbol{F}^{g}$, which are then performed element-wise addition to produce fused hierarchical features $\hat{\boldsymbol{F}}_t$.
Third, following the pSp, 18 mapping networks are trained to extract the learned styles from the hierarchical feature maps. All style vectors are fed into the StyleGAN2 generator to synthesize raw swapped face $\hat{\boldsymbol{I}}_{s \rightarrow t}$. The feature maps of StyleGAN2 are extracted to generate soft mask $\boldsymbol{M}$ simultaneously by FMP. Finally, $\hat{\boldsymbol{I}}_{s \rightarrow t}$ and $\boldsymbol{I}_t$ are blended by $\boldsymbol{M}$ to produce swapped face $\hat{\boldsymbol{I}}_{out}$.

\subsection{Facial Region-Aware Branch} \label{sec:FRA}

\begin{figure}[t!]
	\centering
	\includegraphics[width=0.45\textwidth]{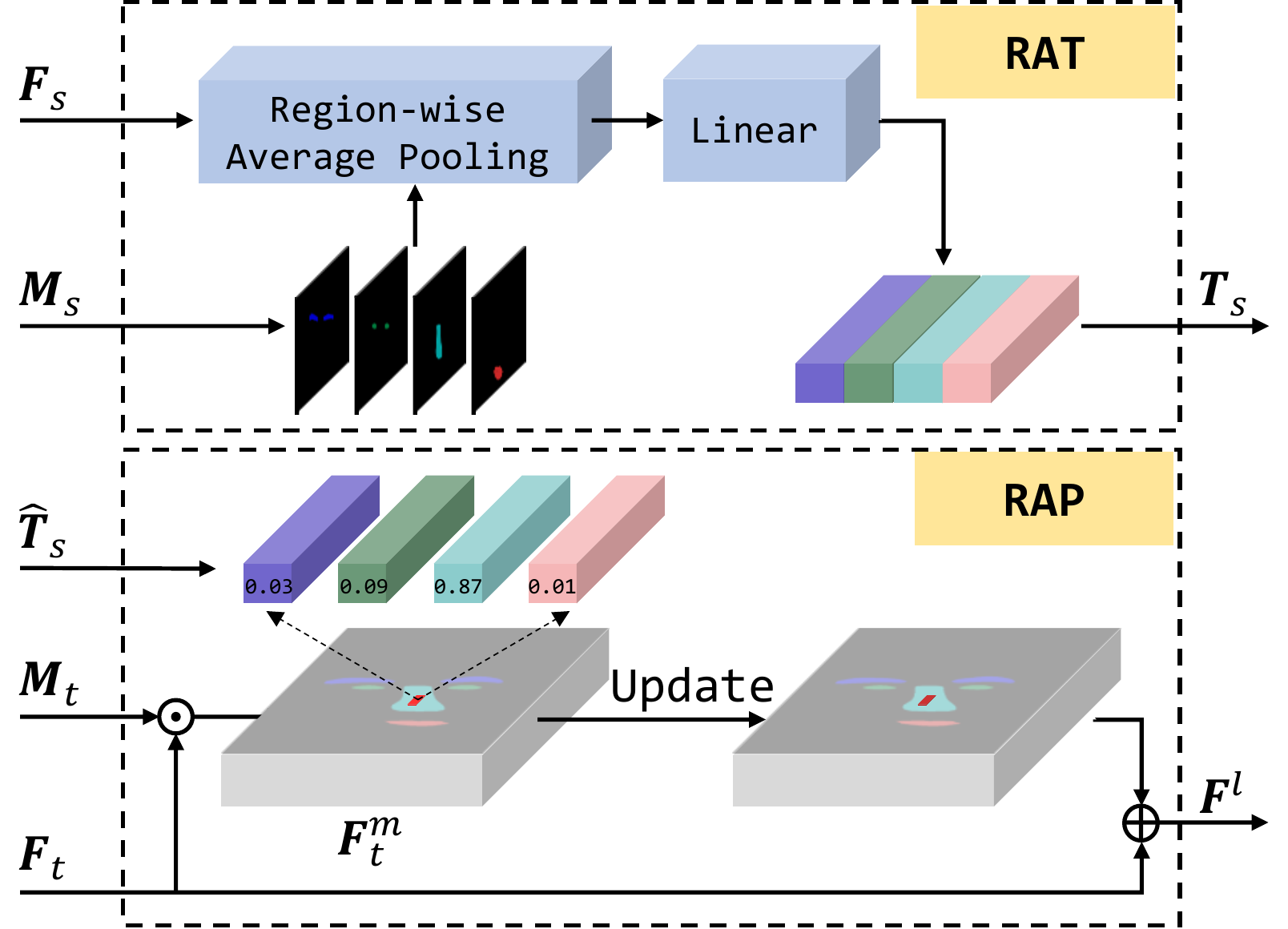}
	\caption{Structure of Region-Aware Identity Tokenizer and  Region-Aware Identity Projector on each scale.}
	\label{fig:tok-proj}
\vspace{-1em}
\end{figure}

\noindent\textbf{Region-Aware Identity Tokenizer.} In order to explicitly model facial features by local identity-relevant regions, \ie, lips, nose, brows, and eyes, we propose a \textit{Region-Aware Identity Tokenizer}. As illustrated in Fig.~\ref{fig:tok-proj}, the purpose of RAT is to convert source face features $\boldsymbol{F}_s$ into compact sets of crucial local identity-relevant tokens ${\boldsymbol{T}_s} \in \mathbb{R}^{N \times L \times 512}$, where $N$ is the number of feature map scales, $L$ is the number of regions. We define three scales and four facial areas, so the $N$ and $L$ are set to 3 and 4. Following the SEAN~\cite{zhu2020sean}, we adopt a region-wise average pooling layer $\Phi$ to obtain the local semantic representations. Specifically, we resize the semantic labels by using bilinear interpolation to match the size of each source feature map. Then, each region's pixel-level features are aggregated and averaged into a corresponding token. A linear layer is followed to embed all hierarchical identity-relevant tokens further.
The tokenizer operation could be represented in the following formula: 
\begin{equation}
\begin{aligned}
\boldsymbol{T}_s^{n} &= \text{Linear}(\Phi(\boldsymbol{F}_s, \boldsymbol{M}_s^{n})),\\
\text{where}~\boldsymbol{M}_s^n &\in \{\boldsymbol{M}_s^{lips}, \boldsymbol{M}_s^{nose}, \boldsymbol{M}_s^{brows}, \boldsymbol{M}_s^{eyes}\}.
\end{aligned}
\end{equation}

\noindent\textbf{Transformer Layers.} The AdaIN-based methods lack feature interaction among crucial local features that cause swapped faces in poor identity consistency. Benefiting from our region-aware mechanism, we introduce the Transformer layer $\mathcal{T}$ to model the interaction between tokens across different scales and semantics, which is built upon the Multi-head Self-Attention (MSA) layer, along with Feed-Forward Network (FFN), Layer Normalization (LN), and Residual Connection (RC) operations. In practice, a reshape operation apply on $\boldsymbol{T}_s$ to combine N and L dimensions: ${\boldsymbol{T}}_s \in \mathbb{R}^{NL \times 512}$. We denote $\boldsymbol{T}_s$ as Query, Key, and Value, respectively. Each attention head is formulated as:
\begin{equation}
\begin{aligned}
\resizebox{.9\hsize}{!}{$ \begin{split}
Attention(\boldsymbol{T}_s) &= \text{Softmax} \left[\frac{\boldsymbol{T}_s \boldsymbol{W}^{Q}\left(\boldsymbol{T}_s \boldsymbol{W}^{K}\right)^{T}}{\sqrt{d_{k}}}\right] \boldsymbol{T}_s \boldsymbol{W}^{V} \\
                  &= \boldsymbol{A} \boldsymbol{T}_s \boldsymbol{W}^{V},
\end{split}
$}
\end{aligned}
\label{equal:att}
\end{equation}
where $\boldsymbol{W}^{Q} \in \mathbb{R}^{d_{m} \times d_{k}}, \boldsymbol{W}^{K} \in \mathbb{R}^{d_{m} \times d_{k}}, \boldsymbol{W}^{V} \in \mathbb{R}^{d_{m} \times d_{v}}$ are parameter matrices for feature projections. $d_{m}$ is the input dimension, while $d_{k}$ and $d_v$ are hidden dimensions of each projection subspace, $\boldsymbol{A} \in \mathbb{R}^{NL \times NL}$ is the attention matrix, which indicates the relation between all tokens. For FFN, which consists of two cascaded linear transformations with a ReLU activation in between:
\begin{equation}
\begin{aligned}
\text{FFN}(\boldsymbol{x})=\max \left(0, \boldsymbol{x} \boldsymbol{W_{1}}+\boldsymbol{b}_{1}\right) \boldsymbol{W_{2}}+\boldsymbol{b}_{2},
\end{aligned}
\end{equation}
where $\boldsymbol{x}$ is the input tokens, $\boldsymbol{W}_{1}$ and $\boldsymbol{W}_{2}$ are weights of two linear layers, and $\boldsymbol{b}_1$ and $\boldsymbol{b}_2$ are corresponding bias. 
The transformed tokens $\hat{\boldsymbol{T}_s}$ is formulated as:
\begin{equation}
\begin{aligned}
\hat{\boldsymbol{T}_{s}} = \boldsymbol{T}_s + [\text{MSA} | \text{FFN}](\text{LN}(\boldsymbol{T}_s)).
\end{aligned}
\end{equation}
Subsequently, each token contains sufficient multi-scale and multi-semantic representation via the Transformer layers.

\noindent\textbf{Region-Aware Identity Projector.} Corresponding to Tokenizer, we need to project the identity-relevant tokens to the target features spatially while considering the misaligned attributes between source and target faces, \eg, gaze and expression. Unlike SEAN~\cite{zhu2020sean} that replaces the style of edited source region by reference style faces, we devise a \textit{Region-Aware Identity Projector} to adaptively transfer identity information to the target face and keep its attributes unchanged. As shown in Fig.~\ref{fig:tok-proj}, the masked target feature map $\boldsymbol{F}_t^m$ is updated by combining weighted $\hat{\boldsymbol{T}}_s$ to refine the $\boldsymbol{F}_t$ for forming local identity-augmented features $\boldsymbol{F}^l$. Specifically, $\boldsymbol{F}_t^{m}$ is flattened along the height and width dimension: $\boldsymbol{F}_{t}^{m} \in \mathbb{R}^{HW \times 512}$ . Given the flattened feature $\boldsymbol{F}_{t}^{m}$ of each scale as Query, the identity-relevant tokens $\hat{\boldsymbol{T}_s}$ of each scale as Key and Value, the attention matrix $\boldsymbol{A}^{P}$ is computed as Eq.~\ref{equal:att}. Each element of $\boldsymbol{A}_{ij}^{P}$ indicates the relation between each pixel and token. We show the scores between a pixel located on the nose and all four tokens in Fig.~\ref{fig:tok-proj}. As expected, it has the highest value with the token extracted from the source nose region. The identity-relevant tokens are thus linearly transferred to $\boldsymbol{F}_t^{m}$, which are then reshaped to the same size as $\boldsymbol{F}_t$ and further added to $\boldsymbol{F}_t$:

\begin{equation}
\begin{aligned}
\boldsymbol{F}^{l} = \boldsymbol{F}_{t} + \text{RS}(\boldsymbol{A}^{P} \hat{\boldsymbol{T}}_s \boldsymbol{W}^{P}),
\end{aligned}
\end{equation}

where $\boldsymbol{W}^{P}$ is learnable weight, RS is reshape operation
\subsection{Source Feature-Adaptive Branch} \label{sec:SFA}
After the FRA, the crucial local identity-relevant features from the source face have been combined into the target face. However, some global facial representations also affect the identity consistency of swapped faces, \eg, skin wrinkle, the relative distance of facial components. Thus, we design a global \textit{Source Feature-Adaptive} branch that captures global information as a complementary cue to distinguish different identities. As shown in Fig.~\ref{fig:pipeline}, to avoid spatial misalignment between source and target faces, the source feature map with the smallest size first goes through a global averaging pooling (GAP). Then MLPs are followed to further adaptively recombine the global features. Finally, we broadcast the global features as large as three scales and add them to the $\boldsymbol{F}^l$ with the same resolution to obtain integrated target features $\hat{\boldsymbol{F}_{t}}$:
\begin{equation}
\begin{aligned}
\boldsymbol{F}^{g} &= \text{MLPs}(\text{GAP}(\boldsymbol{F}_{s}^{0})),\\
\hat{\boldsymbol{F}_{t}} &= \boldsymbol{F}^{g} + \boldsymbol{F}^{l}.
\end{aligned}
\end{equation}

\subsection{Face Mask Predictor} 
\begin{figure}[t!]
	\centering
	\includegraphics[width=0.45\textwidth]{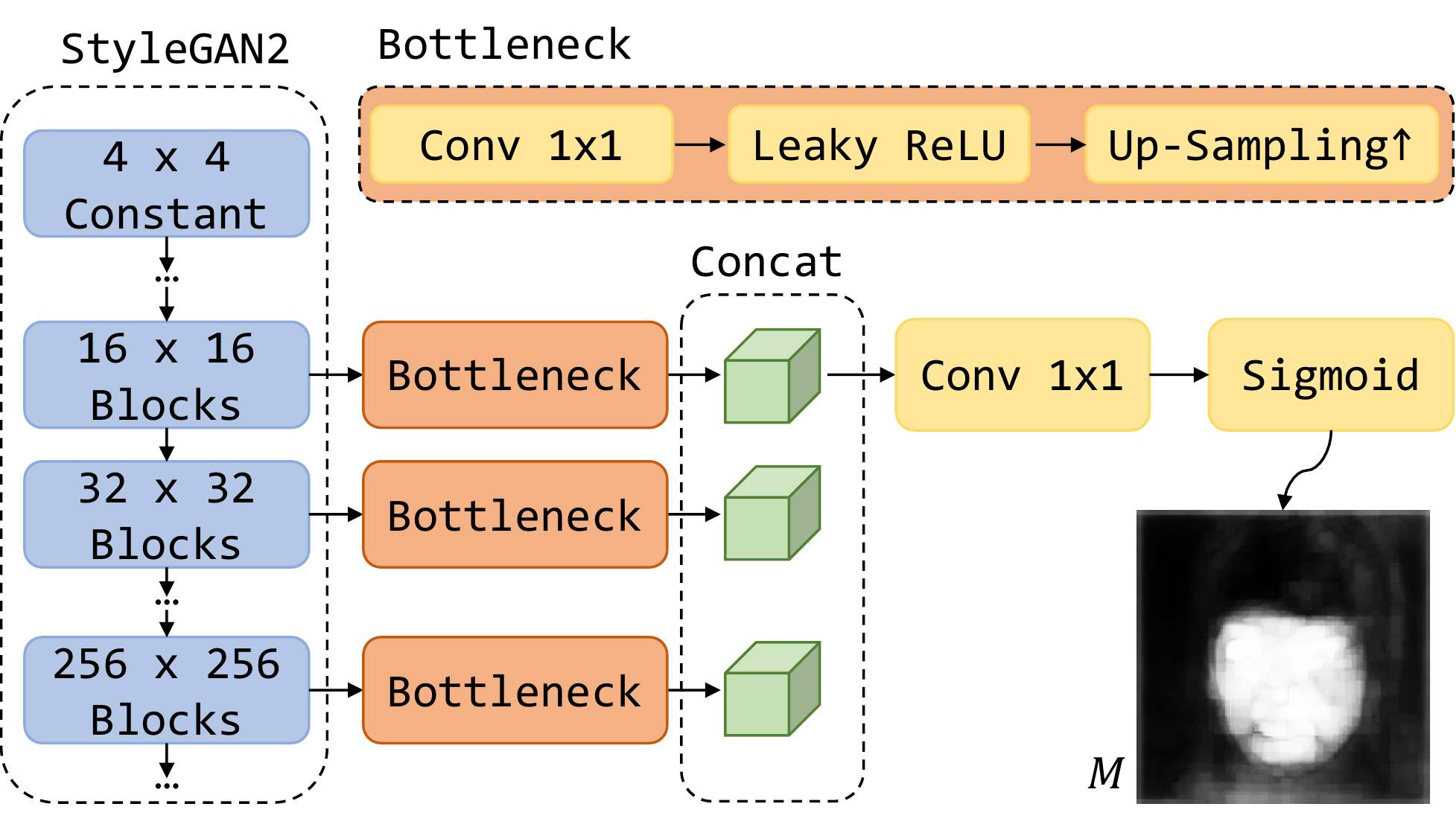}
	\caption{Structure of Face Mask Predictor.}
	\label{fig:mask}
\end{figure}

In order to solve the occlusion and distorted background problem introduced by the GAN inversion framework, MegaFS~\cite{zhu2021one} directly utilize the hard face mask produced by the pre-trained segmentation model for blending, which tends to produce artifacts around the edges and is not computationally-friendly. Instead, inspired by Labels4Free~\cite{abdal2021labels4free}, we make full use of the existing structure. First, the layers of a pre-trained StyleGAN2 already contain rich semantic prior. Second, the identity-consistent constraint could force the mask module to focus on the identity-relevant areas. Thus, we exploit the feature maps of StyleGAN2 to produce soft face masks without specific mask supervision. Specifically, as shown in Fig.~\ref{fig:mask}, we first sample feature maps with a resolution ranging from 16 to 256, then apply a bottleneck on each feature map, which reduces the channel to 32 and upsample the resolution to 256. Finally, the concatenated feature maps are fed to a $1 \times 1$ convolution layer and a sigmoid layer sequentially to produce a single channel soft mask $\boldsymbol{M}$. To generate the swapped face, we blend $\hat{\boldsymbol{I}}_{s \rightarrow t}$ to the target face $\boldsymbol{I}_{t}$ by $\boldsymbol{M}$, formulated as:
\begin{equation}
\begin{aligned}
\hat{\boldsymbol{I}}_{out} = \boldsymbol{M} \odot \hat{\boldsymbol{I}}_{s \rightarrow t} + (1-\boldsymbol{M}) \odot \boldsymbol{I}_t.
\end{aligned}
\end{equation}

\subsection{Objective Functions}
During the training stage of RAFSwap, we adopt \textit{identity loss}, \textit{reconstruction loss}, and \textit{perceptual loss}.

\noindent\textbf{Identity Loss.} A well-trained face recognition model can provide representative identity embeddings. We use cosine similarity to estimate the similarity between the identity embedding of the generated face and the source face, which can be written as:
\begin{equation}
    \begin{aligned}
        \mathcal{L}_{id}=1-\cos (R(\boldsymbol{I}_{s}), R(\hat{\boldsymbol{I}}_{out})),
    \end{aligned}
\end{equation}
where $R(\cdot)$ is a pre-trained ArcFace~\cite{deng2019arcface} network. 

\noindent\textbf{Reconstruction Loss.} If the source and the target faces are from the same identity, the generated face should look the same as the target face. We define a reconstruction loss as pixel-level $\mathcal{L}_{2}$ distances between the target face and the generated face, which can be written as:
\begin{equation}
\mathcal{L}_{rec} = 
\begin{cases}
\left\|\hat{\boldsymbol{I}}_{out} - \boldsymbol{I}_{t}\right\|_2 & \textit{if}~\boldsymbol{I}_t = \boldsymbol{I}_s \\
0 & \textit{otherwise}
\end{cases}.
\end{equation}

\noindent\textbf{Perceptual Loss.} Besides measuring the difference between two faces at the pixel level, we utilize LPIPS~\cite{zhang2018unreasonable} loss to calculate semantic errors between the target and generated faces. It can be written as:

\begin{equation}
    \begin{aligned}
    \mathcal{L}_{p}= \left\|\phi_p(\hat{\boldsymbol{I}}_{out})-\phi_p(\boldsymbol{I}_{t})\right\|_{2},
    \end{aligned}
\end{equation}
where $\phi_p(\cdot)$ represents the pre-trained VGG16 network.

The total loss is the weighted sum of all the above losses:
\begin{equation}
\begin{aligned}
\mathcal{L}_{total}=\lambda_{id} \mathcal{L}_{id}+\lambda_{rec} \mathcal{L}_{rec}+\lambda_{p} \mathcal{L}_{p}.
\end{aligned}
\end{equation}

\section{Experiments}
\subsection{Dataset and Implementation Details}
\noindent\textbf{Dataset.} For face swapping, CelebA-HQ~\cite{karras2017progressive} is a high-quality version of the CelebA~\cite{liu2015deep}, which has 30000 images with 1024 resolution. FaceForensics++~\cite{rossler2019faceforensics++} is a forensic dataset consisting of 1000 video sequences from YouTube.

\noindent\textbf{Implementation Details.} We use CelebA-HQ dataset as the training set, and the values of the loss weights are set to $\lambda_{id}=0.15$, $\lambda_{rec}=1$, $\lambda_{p}=0.8$, respectively. The ratio of the training data with $\boldsymbol{I}_{t} = \boldsymbol{I}_{s}$ and $\boldsymbol{I}_{t} \ne \boldsymbol{I}_{s}$ is set to $1:4$. The input images are resized to $256 \times 256$. During the training, the StyleGAN2 is fixed and the weights of the rest are updated by using Adam optimizer with $\beta_{1}=0.9$, $\beta_{2}=0.999$, and learning rate=$1e^{-4}$. RAFSwap is trained with 50K steps, using 1 Tesla V100 GPU and 8 batch size.

\subsection{Comparison with Previous Methods}
\begin{figure}[t!]
	\centering
	\includegraphics[width=0.45\textwidth]{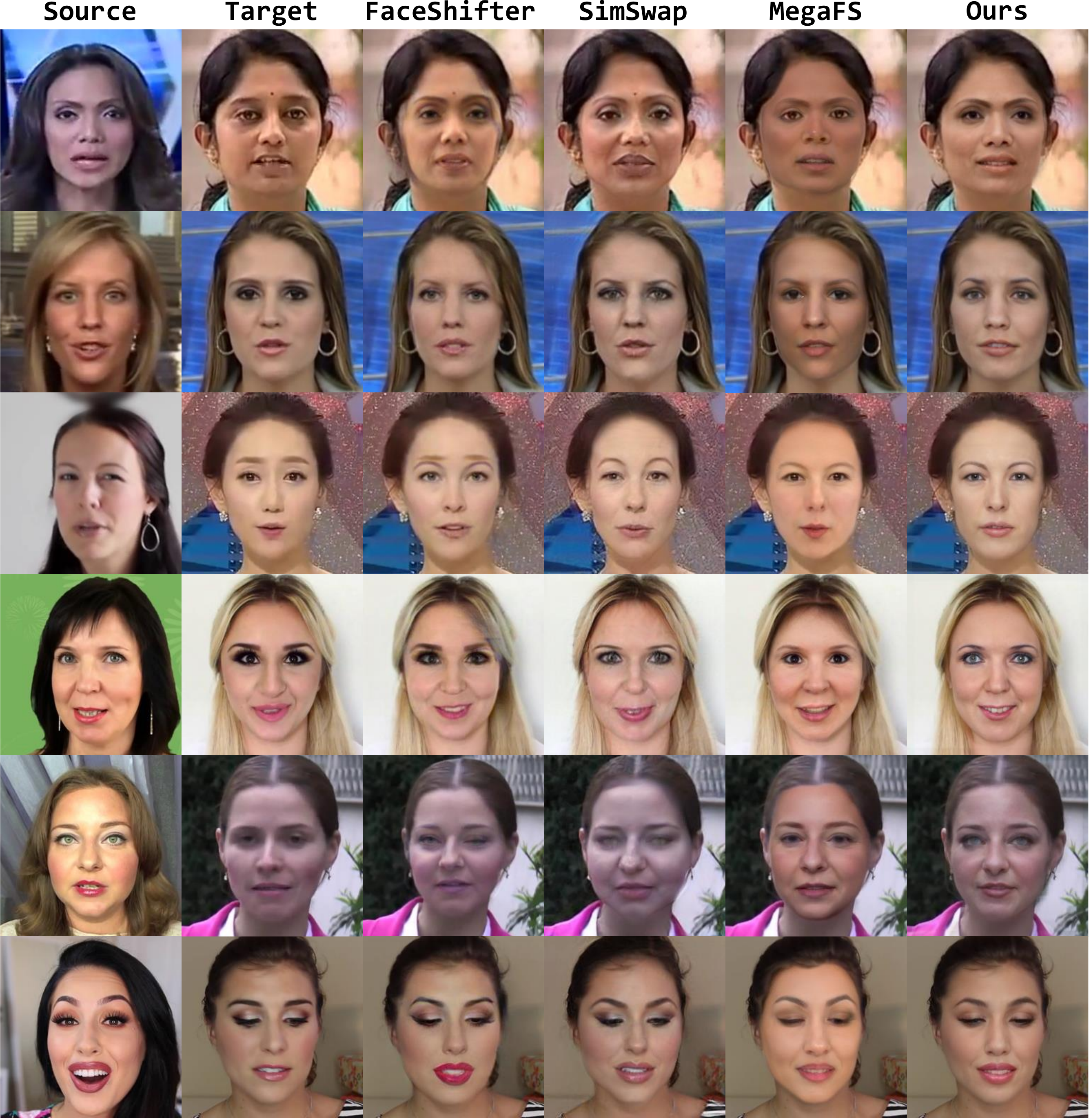}
	\caption{Comparison with FaceShifter~\cite{li2019faceshifter}, SimSwap~\cite{chen2020simswap}, and MegaFS~\cite{zhu2021one} on FaceForensics++~\cite{rossler2019faceforensics++}.}
	\label{fig:cmp_ff++}
\end{figure}

\begin{figure}[t!]
	\centering
	\includegraphics[width=0.40\textwidth]{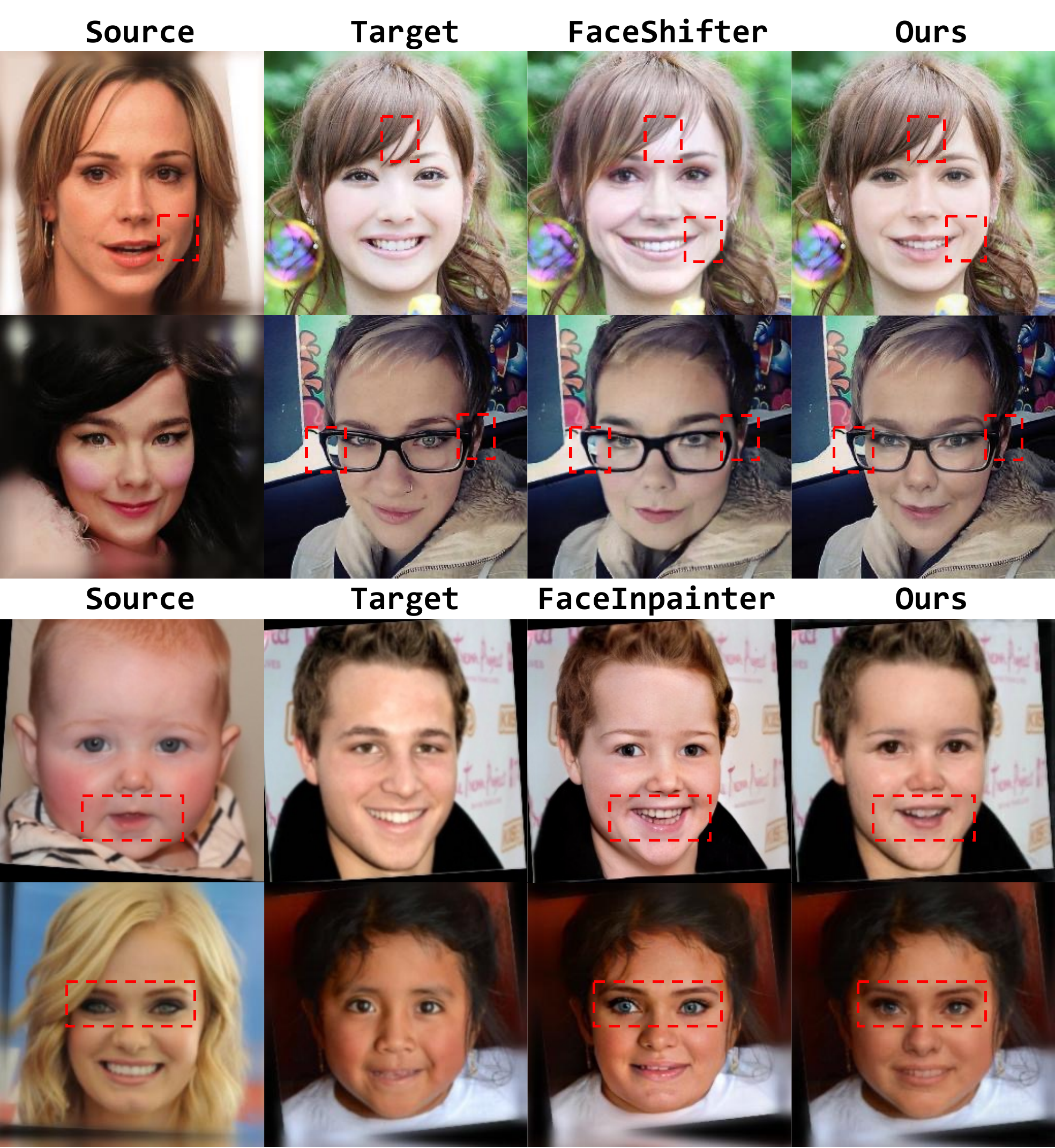}
	\caption{Comparison with FaceShifter~\cite{li2019faceshifter} and FaceInpainter~\cite{li2021faceinpainter}. Images are from official attached results. Please zoom in the red dotted rectangles for a more clear comparison.}
	\label{fig:cmp_fs_occlusion}
\end{figure}

\noindent\textbf{Qualitative Comparison.} We compare our method with FaceShifter~\cite{li2019faceshifter}, SimSwap~\cite{chen2020simswap}, and MegaFS~\cite{zhu2021one} on FaceForensics++. As shown in Fig.~\ref{fig:cmp_ff++}, we show some conditions that are prone to produce artifacts, including face shape, hairstyle, and brows with a large difference between source and target faces. We can see that MegaFS and our method can handle these challenges, but MegaFS could not preserve the attributes of the target face, such as skin color. Besides, our results share eye color with the source face much better than other methods in rows 4 and 5. Furthermore, we compare our method with FaceShifter and FaceInpainter~\cite{li2021faceinpainter} on wild face images. As shown in Fig.~\ref{fig:cmp_fs_occlusion}, benefiting from the well-designed identity integration and flexible soft mask generation modules, our results can well preserve the source identity information, \eg, small mouth, target attributes, \eg, hair color, and handle occlusion cases, \eg, eyeglasses. 

\begin{table}[t!]
\centering
\begin{tabular}{lccc}
\hline
Method         & ID Ret.$\uparrow$ &  Pose$\downarrow$ &  Exp.$\downarrow$  \\ \hline
DeepFakes~\cite{perov2020deepfacelab}     & 88.39          &  4.46             & 3.33   \\
FaceShifter~\cite{li2019faceshifter}    &       90.68         &    2.55    &      2.82     \\ 
SimSwap~\cite{chen2020simswap}    &       89.73        &    \textbf{1.94}    &      \textbf{2.39}     \\ 
MegaFS~\cite{zhu2021one}    &       90.83         &    2.64    &       2.96      \\ \hline
Ours           &      \textbf{96.70}          &      2.53  &        2.92       \\ \hline
\end{tabular}
\caption{Quantitative comparison results on FaceForensics++~\cite{rossler2019faceforensics++}. \textbf{Bold} represent optimal result. The up arrow indicates that the larger the value, the better the model performance, and vice versa.}
\label{tab:ff++}
\end{table}

\begin{figure}[t!]
	\centering
	\includegraphics[width=0.45\textwidth]{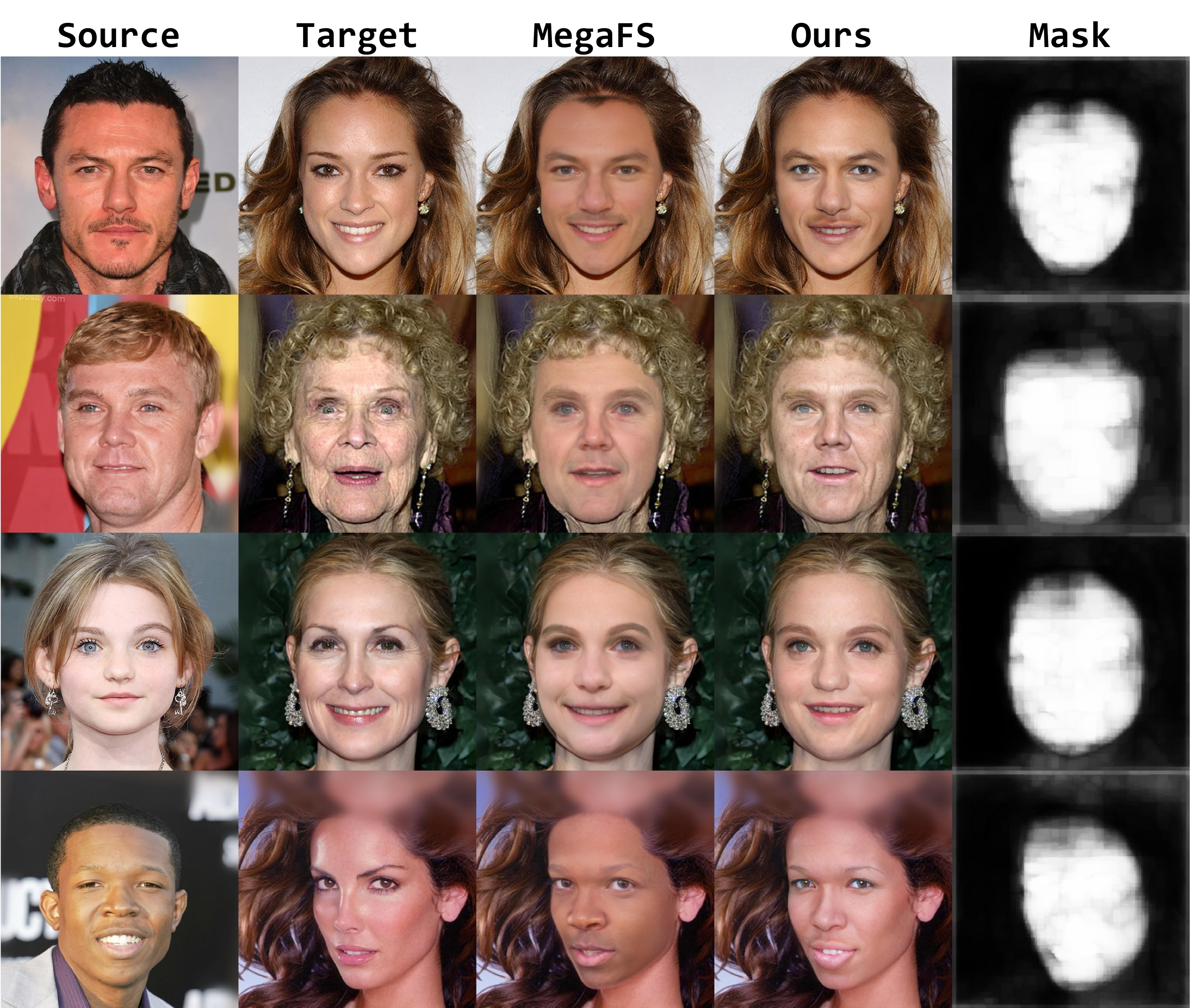}
	\caption{Comparison with MegaFS~\cite{zhu2021one} on CelebA-HQ~\cite{karras2017progressive}. We sample some challenging conditions across significant gaps in gender, age, skin color, and pose.}
	\label{fig:cmp_celeba}
\end{figure}

Since our method can generate high-resolution swapped faces, we compare RAFSwap with MegaFS on CelebA-HQ. As shown in Fig.~\ref{fig:cmp_celeba}, we sample four pairs of significant gaps between gender, age, skin color, and pose. Obviously, our method achieves higher identity-consistent results that share the same local and global representations with source faces, \eg, eye color and skin wrinkle, and faithfully respect the attributes of the target face. Note that our method produces more harmonious fusion results around edges. 

\begin{table}[t!]
\centering
\begin{tabular}{lcccc}
\hline
Method         & ID Sim.$\uparrow$ &  Pose$\downarrow$    &  Exp.$\downarrow$  & FID$\downarrow$ \\ \hline
MegaFS~\cite{zhu2021one}    &       0.4837         &  3.85      &      \textbf{3.13}   & 18.81  \\ \hline
Ours           &      \textbf{0.5232}          &      \textbf{3.77}  &    3.15      & \textbf{13.25}     \\ \hline
\end{tabular}
\caption{Quantitative comparison results on CelebA-HQ~\cite{karras2017progressive}.}
\label{tab:celeba}
\end{table}

\noindent\textbf{Quantitative Comparison.} We follow the experiment settings in MegaFS, which is slightly different from FaceShifter in data preprocessing. Firstly, we sample 10 frames from each video and process them by MTCNN~\cite{zhang2016joint}, resulting in 10K aligned faces. Because some videos display repeated identities and contain multiple faces in one frame, we carefully check the aligned faces and manually categorize all videos into 885 identities. Then, we evaluate the accuracy of identity retrieval (abbrev. ID Ret.), pose, and expression errors (abbrev. Exp.). We apply CosFace~\cite{wang2018cosface} to extract identity embedding and retrieve the closest face by using cosine similarity. A pose estimator~\cite{ruiz2018fine} and 3D facial model~\cite{deng2019accurate} are used to extract pose and expression vectors for pose and expression evaluation. We measure the $\mathcal{L}_{2}$ distances between swapped faces and the corresponding target faces. The comparison results are shown in Tab.~\ref{tab:ff++}, SimSwap preserves better attributes of the target face but a poor identity consistency.
Our method achieves the highest ID retrieval, outperforming MegaFS with a large margin, and the comparable pose and expression errors with FaceShifter. Note that we have omitted comparisons with FaceInpainter quantitatively since the source codes are not publicly available.

For high-resolution swapped results comparison with MegaFS, we randomly sample 100K pairs of the face images in the CelebA-HQ test set. We report ID similarity (abbrev. ID Sim.), pose errors, expression errors, and FID. ID similarity is measured by calculating the cosine similarity of swapped faces and the corresponding source faces. As shown in Tab.~\ref{tab:celeba}, RAFSwap achieves a better performance in ID similarity and pose error than MegaFS but has a higher expression error. Because MegaFS adopts landmark loss, which produces the swapped face that faithfully respects the mouth shape of the target but leads to low identity-consistent with the source. Besides, the lower FID indicates that our method could generate more realistic images.

\noindent\textbf{Human Study.} We conduct a human study to evaluate the performance of each method. Corresponding to two challenges, we let the users select: \romannumeral1) the one that has the most similar identity with the source face and shares the most similar attributes with the target face; \romannumeral2) the most high-quality one. For each user, we randomly sample 20 pairs from the 1000 FaceForensics++ videos without duplication. The results reported in Tab.~\ref{tab:hs} are based on the answers from 50 users, showing that our method significantly surpasses the other three methods.

\begin{table}[t!]
\centering
\begin{tabular}{lccc}
\hline
Method         & Identity-Perception$\uparrow$ &  Quality$\uparrow$              \\ \hline
DeepFakses~\cite{perov2020deepfacelab}    &      0.07          & 0.05  \\
FaceShifter~\cite{li2019faceshifter}    &       0.16        &    0.13         \\ 
SimSwap~\cite{chen2020simswap}       &     0.13           &    0.09                \\
MegaFS~\cite{zhu2021one}    &       0.15        &    0.16                  \\ \hline
Ours           &      \textbf{0.49}          &  \textbf{0.57}     \\ \hline
\end{tabular}
\caption{Human Study results on FaceForensics++~\cite{rossler2019faceforensics++}.}
\label{tab:hs}
\end{table}

\begin{figure}[t!]
	\centering
	\includegraphics[width=0.45\textwidth]{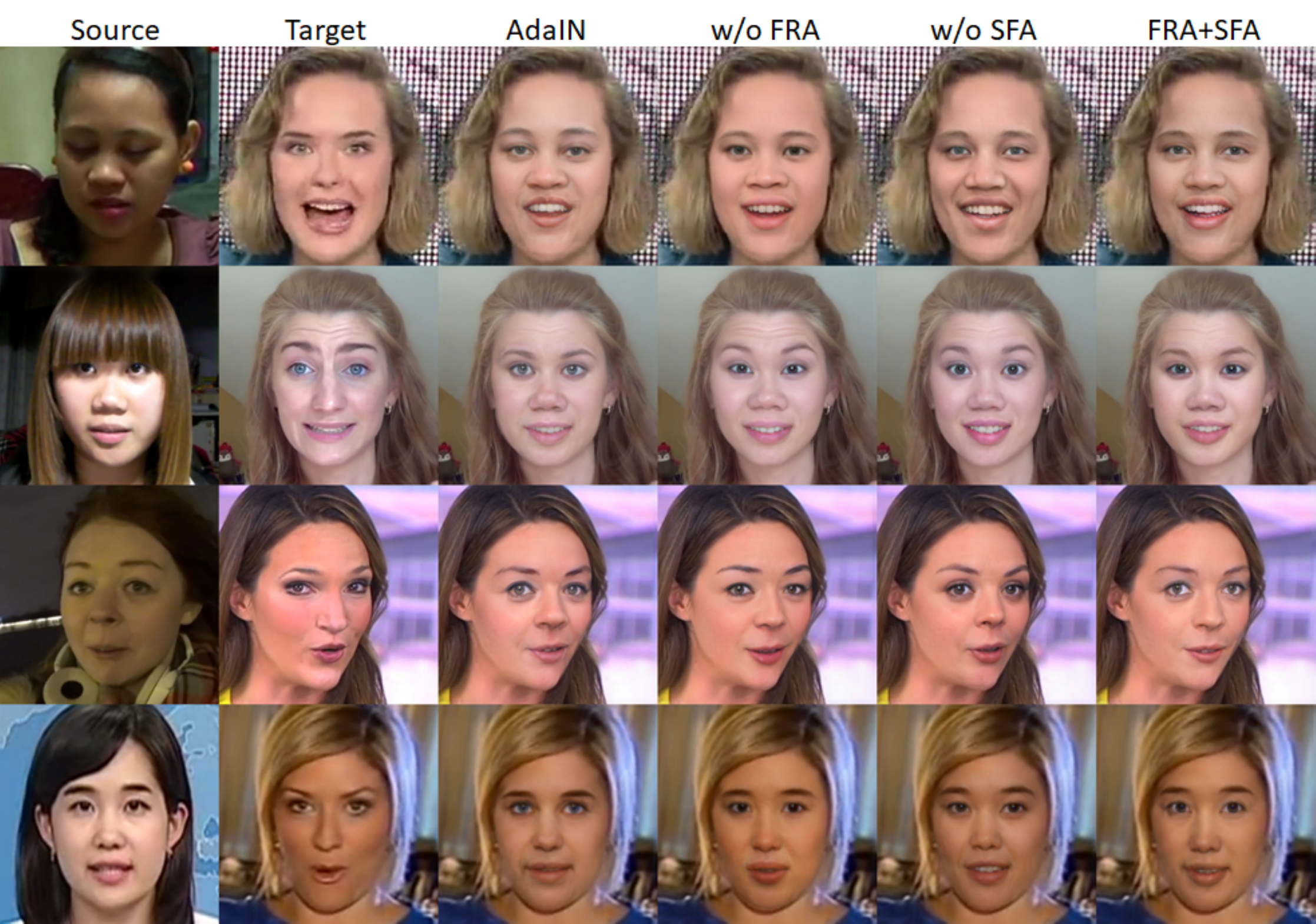}
	\caption{Qualitative results of the ablation study. Our full model obtains better results than other variants.}
	\label{fig:ablation}
\end{figure}

\subsection{Ablation Study and Applications}

\noindent\textbf{Feature Fusion module.} To verify that the combination of FRA and SFA is superior to AdaIN in the GAN inversion framework, we conduct qualitative and quantitative experiments. Specifically, we modify an AdaIN-based baseline that injects the identity vector into hierarchical feature maps. As shown in rows 2 and 4 of Fig.~\ref{fig:ablation}, the third column images produced by AdaIN could not preserve the facial identity details of the source face. Due to the global AdaIN operation and Fixed StyleGAN2 generator, this baseline could not adaptively maintain the detailed identity features and tend to express the general representation. For comparison, our method could generate more identity-consistent faces thanks to the well-designed feature integration module.
Besides, comparing rows 1 and 5 in Tab.~\ref{tab:ablation_1}, our method markedly improves by 4.22 on ID Ret$.$ over the AdaIN-based baseline. As a cost, an extra BiseNet of 13.3M requires 4 hours of training in CelebA-HQ. Furthermore, we analyze the necessity of FRA and SFA branches. As shown in Fig.~\ref{fig:ablation}, the isolated SFA exhibits a shortage of identity performance while the isolated FRA has the better identity performance but suffers from source facial texture mismatch. When both FRA and SFA are used, the generated faces preserve the local identity-relevant features and the global facial details of the source face. 
The quantitative experiments in Tab.~\ref{tab:ablation_1} consistently demonstrate the effectiveness of each component and the superiority of our module. 

\begin{figure}[t!]
	\centering
	\includegraphics[width=0.35\textwidth]{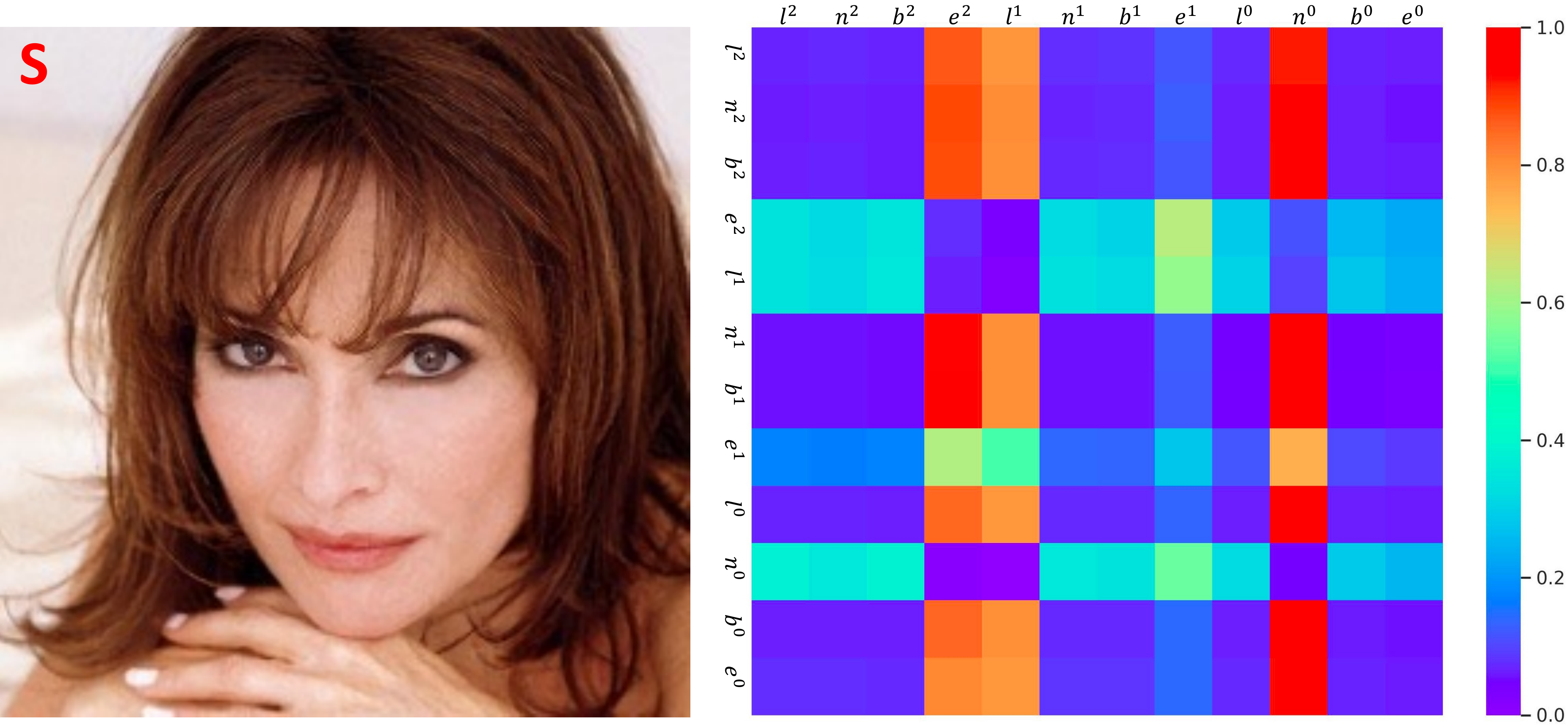}
	\caption{Attention visualization for a source face. Symbols $l, n, b, e$ means lips, nose, brows, eyes, respectively. Superscripts $0, 1, 2$ represent small, medium, and large scales, respectively.}
	\label{fig:attvis}
\vspace{-1em}
\end{figure}

\noindent\textbf{Attention Structure.} To verify the strong ability of the Transformer to capture token interactions, we conduct a quantitative experiment. Specially, we modify a comparative version that replaces a Transformer layer with a Nonlocal layer~\cite{wang2018non}. As shown in the first three rows of Tab.~\ref{tab:ablation_2}, one Transformer layer improves the performance, while one Nonlocal layer could not sufficiently model the token interaction and cause slight performance degradation. Besides, to evaluate the effect of layer numbers, we conduct a controlled experiment. As shown in the last three rows of Tab.~\ref{tab:ablation_2}, as the number of layers increases, the performance does not improve significantly. To balance the performance and calculation, we employ the Transformer with one layer and eight heads experimentally. Besides, we visualize one attention head for a source face. As shown in Fig.~\ref{fig:attvis}, the attention map indicates that the Transformer concentrate on different semantic regions across different scales, \ie, the tokens from the large scale focus on eyes, the medium focus on lips, while the small focus on nose. 
Notably, the brows do not receive much attention due to the small areas and overlapped receptive fields with eyes.

\begin{table}[t!]
\begin{tabular}
{C{22} C{14} C{14} C{14} C{38} C{24} C{24}}
\hline
AdaIN   & FRA  & SFA & FMP  & ID Ret.$\uparrow$   &  Pose$\downarrow$ &  Exp.$\downarrow$  \\ \hline
\cmark  & \xmarkg   &\xmarkg &\xmarkg  &92.48    & 2.60   &  2.98\\ \hline
\xmarkg & \cmark   & \xmarkg &\xmarkg &96.63          & 2.58          & 2.94      \\ 
\xmarkg & \xmarkg &\cmark  &\xmarkg       &93.68           & 2.61          & 3.06  \\ 
\xmarkg &\cmark &\cmark &\xmarkg &   96.69  & 2.54           & 2.94      \\
\xmarkg &\cmark &\cmark &\cmark &\textbf{96.70}  & \textbf{2.53}           & \textbf{2.92}   \\
\hline
\end{tabular}
\caption{Quantitative ablation study of RAFSwap with different proposed components on FaceForensics++~\cite{rossler2019faceforensics++}.}
\label{tab:ablation_1}
\end{table}

\begin{table}[t!]
\centering
\begin{tabular}{lccc}
\hline
Method         & ID Ret.$\uparrow$ &  Pose$\downarrow$ &  Exp.$\downarrow$  \\ \hline
+ Nonlocal~\cite{wang2018non}       &      96.50         &    2.63    &     3.03      \\
+ Tr-0    &       96.62         &    2.60    &      3.01     \\ 
+ Tr-1    &       96.70         &    2.53   &       2.92      \\
+ Tr-2           &      96.71       &      \textbf{2.51}  &     2.94    \\
+ Tr-3           &      \textbf{96.73}          &      2.54  &        \textbf{2.90}  \\
\hline
\end{tabular}
\caption{Quantitative ablation study of RAFSwap with different attention components on FaceForensics++~\cite{rossler2019faceforensics++}.}
\label{tab:ablation_2}
\end{table}

\begin{table}[t!]
\centering
\begin{tabular}
{ccccc}
\hline
\multirow{2}{*}{Method}    & CPU  & GPU  & Params  & Flops  \\
& (s)$\downarrow$   & (ms)$\downarrow$  &(M)$\downarrow$  &(G)$\downarrow$  \\
\hline
LADN~\cite{gu2019ladn}          & 8.70        &  26.8     & 26.99  & 175.77 \\
PSGAN~\cite{jiang2020psgan}         & 8.45  &  128.9      & \textbf{12.61}   & 91.02   \\ \hline
Ours          &   \textbf{0.283}  &  \textbf{9.3}   &  13.60   & \textbf{71.39}  \\ \hline
\end{tabular}
\caption{Efficiency evaluation on makeup transfer. FPS is evaluated on a single Tesla V100.}
\label{tab:makeup}
\end{table}

\begin{figure}[t!]
	\centering
	\includegraphics[width=0.43\textwidth]{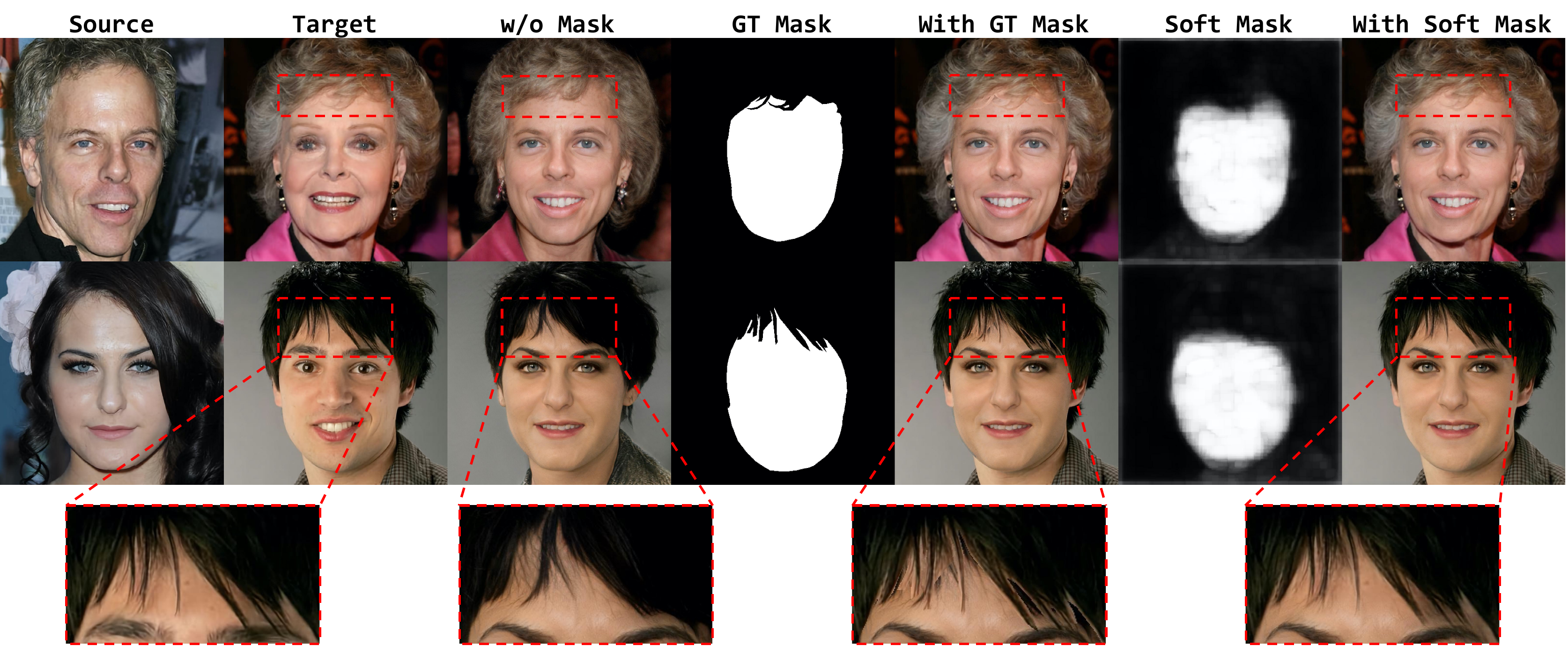}
	\caption{Qualitative results for FMP. We zoom in the red dotted rectangles of the second sample for more clear comparison.}
	\label{fig:abla_mask}
\vspace{-1em}
\end{figure}

\noindent\textbf{Face Mask Predictor.} 
To demonstrate the effectiveness of FMP, we provide two qualitative comparisons. As shown in Fig.~\ref{fig:abla_mask}, without the guidance of the face mask, our method could not keep some attributes unchanged, \eg, background. Applying the hard ground truth mask on raw swapped faces produces excessive information and unnatural edges, especially on the bangs area. In comparison, our full model with the soft mask module achieves more harmonious fusion faces. FMP also brings improvement quantitatively, as shown in the last two rows of Tab.~\ref{tab:ablation_1}.

\begin{figure}[t!]
	\centering
	\includegraphics[width=0.4\textwidth]{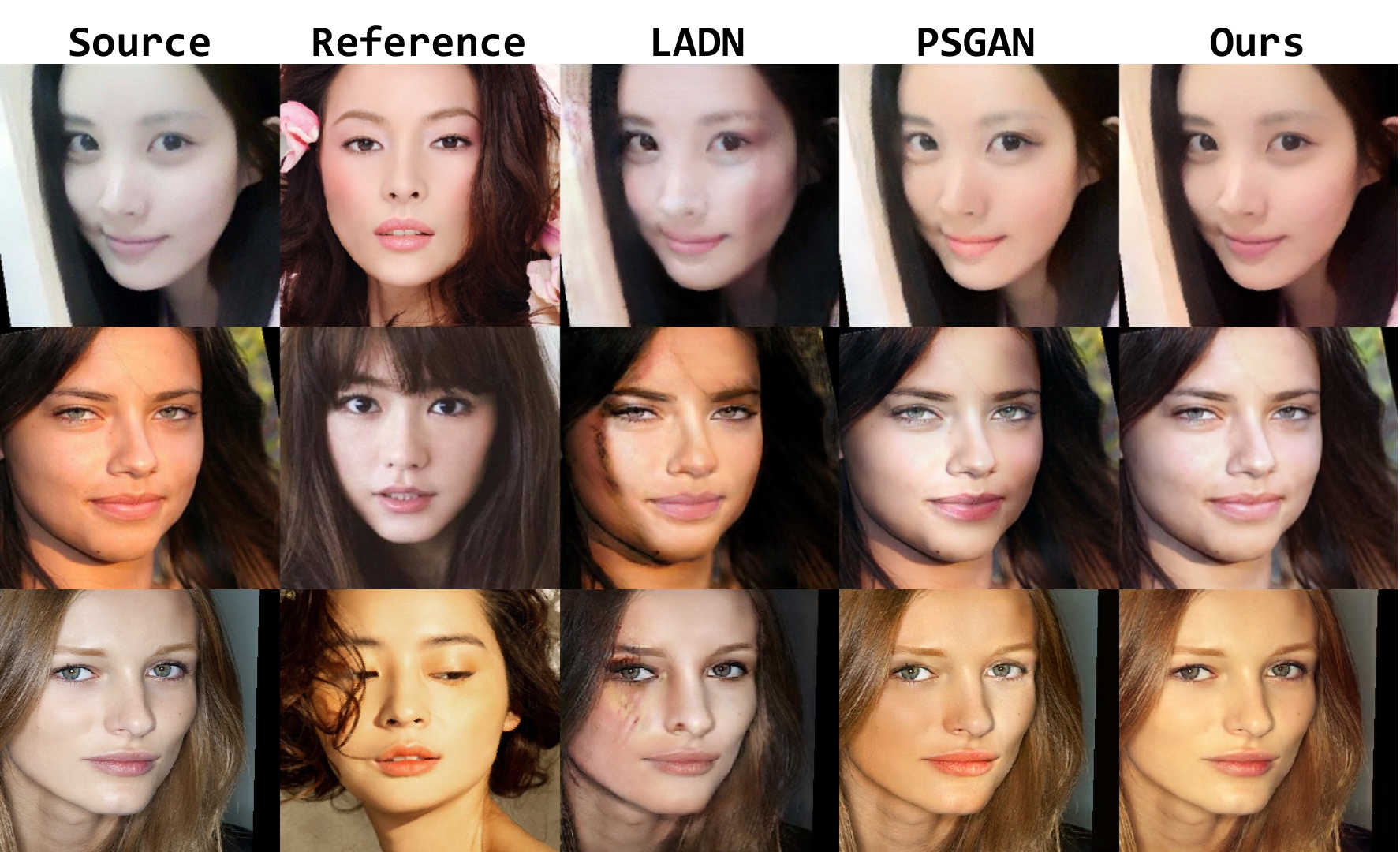}
	\caption{Comparison with SOTA makeup transfer methods on Makeup-Wild dataset~\cite{jiang2020psgan}.}
	\label{fig:makeup}
\vspace{-1em}
\end{figure}

\noindent\textbf{Expanded Application of FRA.} We further apply our FRA branch in makeup transfer. Specifically, we adopt PSGAN~\cite{jiang2020psgan} as the baseline. For a fair comparison, we only replace the AMM module of PSGAN with the FRA branch. As shown in Fig.~\ref{fig:makeup}, compared with LADN~\cite{gu2019ladn} and PSGAN, our method precisely transfers makeup colors with realistic results, where the identity and light on the source face are well preserved. Besides, we compare their running efficiency. The results are shown in Tab.~\ref{tab:makeup}. Our method is more than ten times faster than PSGAN on GPU. The expanded experiment demonstrates that FRA could also handle texture and color feature transfer due to the flexible token mechanism and sufficient feature interaction.

\section{Conclusion and Future Work}
In this paper, we propose a novel RAFSwap built on the GAN inversion to generate high-resolution and identity-consistent swapped faces. Specifically, FRA integrates the identity-relevant local features into the target face, and SFA provides complementary identity-relevant details globally. Furthermore, FMP incorporated with StyleGAN2 is proposed to preserve the background and occlusions of the target unsupervisedly. Extensive experiments demonstrate the superiority of our approach over other SOTA methods.

Due to the limitation of the training dataset, inversion-based methods fail to handle out-range cases, \ie, faces with various perspectives. We will further incorporate prior knowledge to improve the practicability of our method.

\section{Acknowledgments}
This work is supported by the National
Natural Science Foundation of China (NSFC) under Grant
No. 61836015.

{\small
\bibliographystyle{ieee_fullname}
\bibliography{PaperForArxiv}
}

\end{document}